\pdfoutput=1
\documentclass[11pt]{article}

\usepackage{acl}
\usepackage{times}
\usepackage{latexsym}
\usepackage{algorithm}
\usepackage{algpseudocode}
\usepackage{amsmath}
\usepackage{amssymb}
\usepackage{amsmath}
\usepackage{arydshln}
\usepackage{ascii}
\usepackage{bm}
\usepackage{booktabs}
\usepackage{colortbl}
\usepackage{color}
\usepackage{CJK}
\usepackage{dsfont}
\usepackage{enumitem}
\usepackage{forest}
\usepackage{graphics}
\usepackage{graphicx}
\usepackage{latexsym}
\usepackage{multirow}
\usepackage{mwe}
\usepackage{makecell}
\usepackage{microtype}
\usepackage{subfig}
\usepackage{subcaption}
\usepackage{times}
\usepackage{tikz}
\usepackage{tabularx}
\usepackage{url}
\usepackage{wrapfig}
\usepackage{tcolorbox}
\usepackage{xcolor}

\definecolor{sred}{RGB}{203, 64, 46}
\definecolor{sblue}{RGB}{44, 73, 135}
\definecolor{sgreen}{RGB}{37, 100, 28}

\usepackage[T1]{fontenc}
\usepackage[utf8]{inputenc}

\usepackage{microtype}

\usepackage{inconsolata}

\title{UltraLink\includegraphics[width=1.4em]{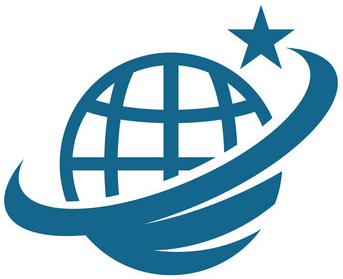}: An Open-Source Knowledge-Enhanced\\
Multilingual Supervised Fine-tuning Dataset}

\author{Haoyu Wang$^{*2}$ \
  Shuo Wang$^{*\dag1}$ \
  Yukun Yan\thanks{\ \ Equal contribution.}$^{1}$ \
  Xujia Wang$^{1}$ \\
  \textbf{Zhiyu Yang}$^{3}$ \
  \textbf{Yuzhuang Xu}$^{1}$ \
  \textbf{Zhenghao Liu}$^{4}$ \
  \textbf{Liner Yang}$^3$ \\
  \textbf{Ning Ding}$^{1}$ \
  \textbf{Xu Han}$^{1}$ \
  \textbf{Zhiyuan Liu}$^{\dag1}$ \
  \textbf{Maosong Sun}\thanks{\ \ Corresponding authors.}$^{1}$ \\
  $^1$Tsinghua University \quad $^2$Beijing University of Posts and Telecommunications \\
  $^3$Beijing Language and Culture University \quad $^4$Northeastern University, China \\
  }
\begin{document}
\maketitle
\begin{abstract}
Open-source large language models (LLMs) have gained significant strength across diverse fields. Nevertheless, the majority of studies primarily concentrate on English, with only limited exploration into the realm of multilingual abilities. In this work, we therefore construct an open-source multilingual supervised fine-tuning dataset. Different from previous works that simply translate English instructions, we consider both the language-specific and language-agnostic abilities of LLMs. Firstly, we introduce a knowledge-grounded data augmentation approach to elicit more language-specific knowledge of LLMs, improving their ability to serve users from different countries. Moreover, we find modern LLMs possess strong cross-lingual transfer capabilities, thus repeatedly learning identical content in various languages is not necessary. Consequently, we can substantially prune the language-agnostic supervised fine-tuning (SFT) data without any performance degradation, making multilingual SFT more efficient. The resulting UltraLink dataset comprises approximately 1 million samples across five languages (i.e., En, Zh, Ru, Fr, Es), and the proposed data construction method can be easily extended to other languages. UltraLink-LM, which is trained on UltraLink, outperforms several representative baselines across many tasks.\footnote{\ \ Both UltraLink and UltraLink-LM will be publicly available at \url{https://github.com/OpenBMB/UltraLink}.}
\end{abstract}

\section{Introduction}

Thanks to the collaborative efforts of the active large language models (LLMs) community, open-source LLMs are becoming increasingly powerful~\cite{llama,llama2,jiang2023mistral}, even outperforming some representative closed-source counterparts~\cite{openai2023gpt4,anil2023palm} in some specific tasks~\cite{wei2023magicoder}.
These accomplishments are closely related to the contribution of open-source supervised fine-tuning (SFT) data~\cite{ding-etal-2023-enhancing,anand2023gpt4all,peng2023instruction,wang-etal-2023-self-instruct,kim-etal-2023-soda,xu-etal-2023-baize}, which plays a pivotal role in eliciting the instruction-following ability of LLMs and aligning the model behavior with human preferences. Nevertheless, the focus of existing works is primarily on the construction of English SFT data, resulting in a comparatively limited availability of multilingual SFT resources.

\begin{figure}[t]
    \centering
    \includegraphics[width=0.8\linewidth]{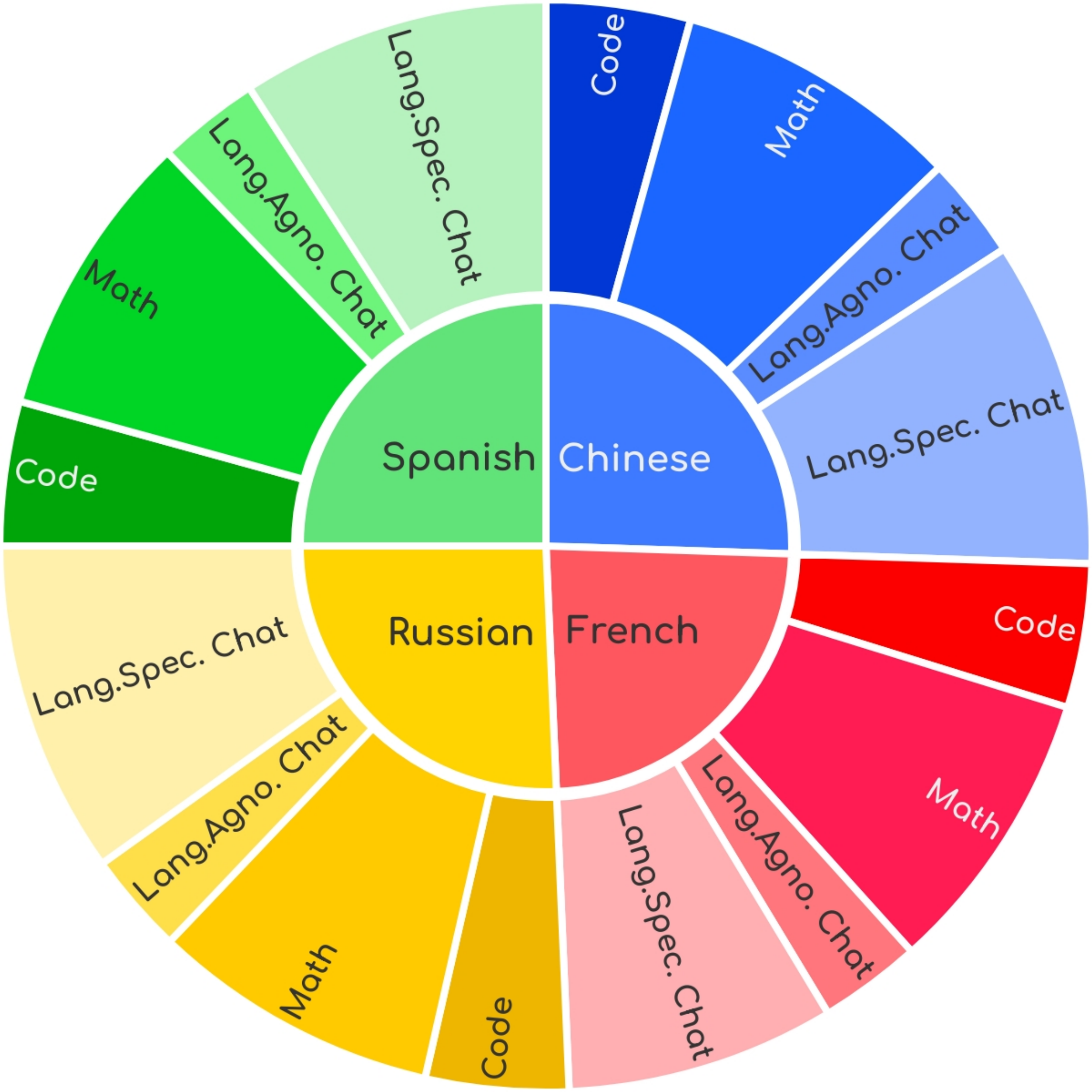}
    \caption{To equip large language models with not only language-specific knowledge but also language-agnostic expertise, we construct the UltraLink dataset for multilingual SFT. For each language, UltraLink consists of four subsets, encompassing chat data with language-specific content, chat data with language-agnostic content, math data, and code data.}
    \label{fig:data_distribution}
\end{figure}

To mitigate the challenge of data scarcity, some researchers suggest translating English SFT data into multiple languages. \citet{okapi} utilize ChatGPT\footnote{\url{https://chat.openai.com}} to translate the two essential components, instructions and responses, from Alpaca-style~\cite{alpaca} English data to other languages. \citet{phoenix} propose to translate both the Alpaca and the ShareGPT\footnote{\url{https://sharegpt.com}} data. While directly translating English SFT data can effectively support multilingual SFT, there are still two major drawbacks associated with this approach:

\begin{itemize}
    \item {\em Low cultural diversity and imprecise translations caused by cultural differences}: translation of English data may not adequately encompass topics specific to non-English regions (e.g., subjects related to Russian culinary culture), leading to a deficiency in language-specific knowledge for LLMs.
    Moreover, for certain instructions (e.g., \texttt{what are the most important holidays of the year?}), the answers vary in different cultural backgrounds, so directly translating all English conversations may result in numerous distorted translations.
    \item {\em Linearly increased data volume}: the total volume of translated SFT data linearly increases with the number of languages. However, the translations across different languages are semantically equivalent, making the model repeatedly learn the same content.  
\end{itemize}

We believe that a good multilingual LLM should not only possess language-specific knowledge but also be equipped with language-agnostic skills. Figure~\ref{fig:lang-example} gives an example of the two types of instructions. We thus propose a new approach to better construct multilingual SFT data, applicable to any language. Compared to conversation translation~\cite{okapi,phoenix}, our advantages can be illustrated as follows:

\begin{itemize}
    \item {\em Higher cultural diversity and less distorted translations}: for language-specific data, we propose a knowledge-grounded data augmentation method. Concretely, Wikipedia is employed\footnote{\url{https://www.wikipedia.org}} as a knowledge base for each language to provide more language-specific contexts.
    For language-agnostic chat data (e.g., the second example in Figure~\ref{fig:lang-example}), we propose a two-stage translation mechanism. Given high-quality English SFT data, we first filter out the conversations that are specific to certain regions. Then we translate the remaining language-agnostic data.
    \item {\em Pruned data volume}: for language-agnostic skills like math reasoning and code generation, through our experiments, we find that it is unnecessary for the model to repeatedly learn identical problems, thanks to the strong cross-lingual transfer capabilities of modern LLMs. We can thus significantly prune the amount of math and code SFT data for non-English languages without compromising the model performance.
\end{itemize}

\begin{figure}[t]
\begin{tcolorbox}[colback=blue!2,colframe=blue!50!black]
\small
\textbf{1. Language-Specific Instructions}\\
\texttt{What are some common tea traditions or etiquette observed in England?}\\
\textbf{2. Language-Agnostic Instructions}\\
\texttt{How do you approach learning a new skill or acquiring knowledge, and what strategies have you found to be effective in your learning process?}
\end{tcolorbox}
\caption{Examples of instructions with language-specific and language-agnostic content.}
\label{fig:lang-example}
\end{figure}

We apply the aforementioned approach to four non-English languages, including Chinese, Russian, French, and Spanish. Note that our method can also be easily extended to other languages. Finally, we train an SFT LLM on the proposed UltraLink dataset, which outperforms several representative open-source multilingual LLMs, demonstrating the effectiveness of our dataset.

\section{Data Curation}
\begin{figure*}[t]
    \centering
    \includegraphics[width=0.99\linewidth]{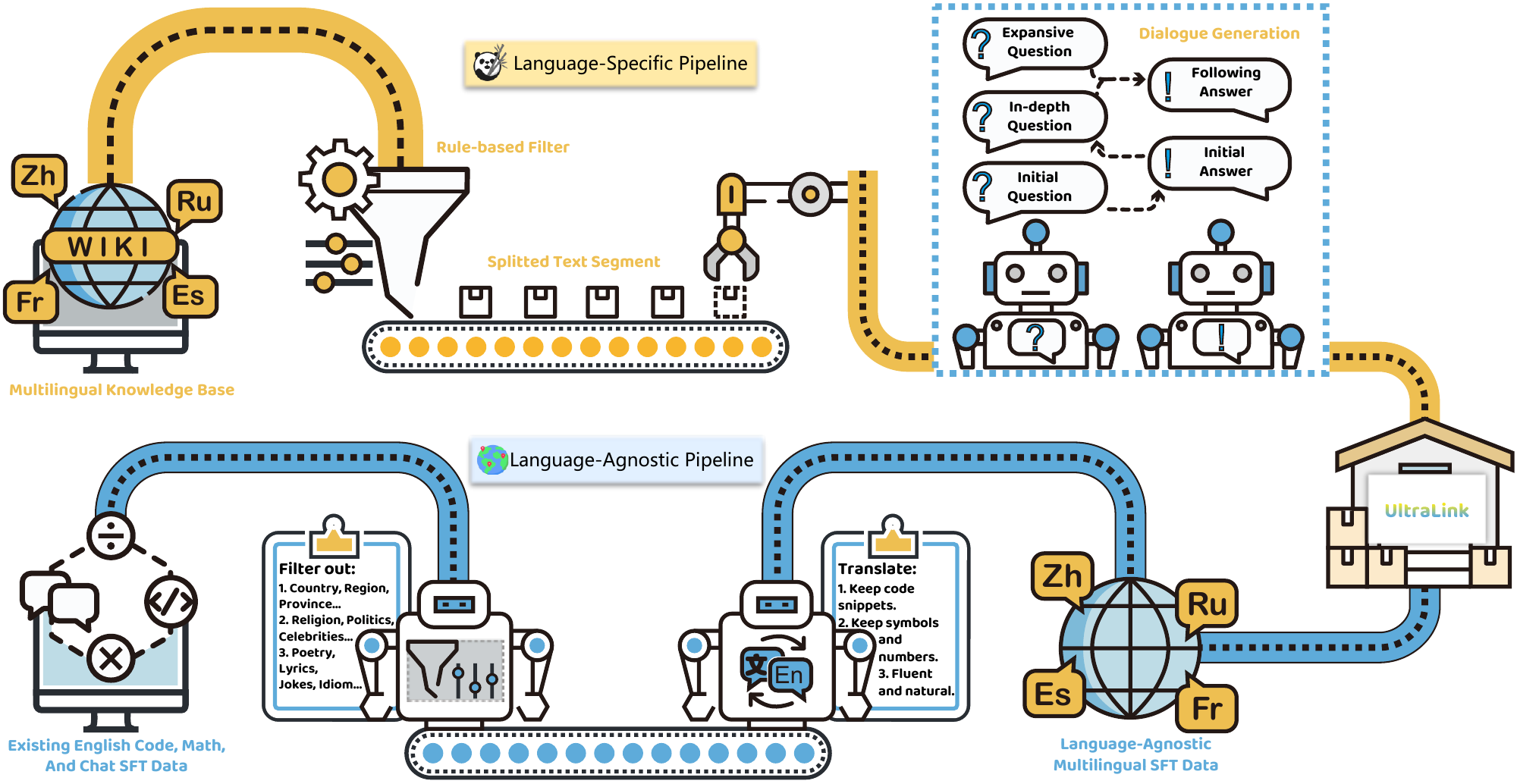}
    \caption{The proposed data augmentation method consists of two pipelines. The upper pipeline illustrates the generation of language-specific chat data. Dialogues are generated by LLMs, conditioning on language-specific knowledge extracted from Wikipedia. The language-agnostic pipeline aims to leverage existing high-quality English SFT data, using a two-stage translation mechanism to mitigate translation errors stemming from cultural differences.}
    \label{fig: flow diagram}
\end{figure*}

Automatically generating SFT data is now an important research topic for LLMs~\cite{alpaca,wang-etal-2023-self-instruct,ding-etal-2023-enhancing}. For multilingual SFT, it is crucial to consider the influence of cultural diversity on language-specific data, while also integrating language-agnostic universal data that is related to the general ability of LLMs (i.e., math reasoning). In this work, we propose a data construction framework consisting of two pipelines, as shown in Figure~\ref{fig: flow diagram}. 

\subsection{Language-Specific Data Curation}

The cultures around the world are vibrant and diverse, reflecting the lifestyles and perspectives of people from various countries and regions. To better cater to diverse users, the cultural diversity of multilingual LLMs should be improved. In this aspect, we propose a knowledge-grounded data augmentation method, leveraging language-specific knowledge bases to provide intricate and varied cultural backgrounds. Our method mainly contains two steps: (1) preparing and sampling knowledge from knowledge bases as cultural backgrounds, and (2) steering LLMs to generate informative conversations given the provided cultural backgrounds.

\subsubsection{Knowledge Preparation}
For each language, we utilize Wikipedia dumps\footnote{\url{https://dumps.wikimedia.org}} as the knowledge base, encompassing a diverse array of topics closely related to the respective culture. We first use an open-source extraction toolkit\footnote{\url{https://github.com/attardi/wikiextractor}} to preprocess the raw dumps and get text descriptions for each entry. Then we use the language identification model provided by \texttt{fastText}~\cite{fasttext} to remove contents that are not in the expected language. For Chinese, we also use \texttt{OpenCC}\footnote{\url{https://github.com/BYVoid/OpenCC}} to convert traditional Chinese texts into simplified Chinese. Finally, we filter out documents that are shorter than 1K tokens or longer than 10K tokens. The number of tokens is calculated by \texttt{tiktoken}\footnote{\url{https://github.com/openai/tiktoken}}.

Given that most LLMs have a limited context length, we divide the whole text into segments whose lengths are between 1K and 2K. We do not split whole sentences when performing text segmentation.
The preprocessed texts are used as contexts for the following dialogue generation procedure.

\subsubsection{Dialogue Generation}

To automatically generate multi-turn dialogues, we designed a question generator and an answer generator, which are both based on GPT-3.5. 
When generating the dialogue, both the question and answer generators are conditioned on a provided text segment as the cultural background.
The used prompts can be divided into four parts: system prompt, principles, cultural background, and dialogue history. The prompt structure is shown in Figure~\ref{fig:prompt-structure}.
The system prompt is used to describe the task (i.e., \texttt{generating the initial question}). The principles provide some detailed suggestions for the LLM, which are found important for improving the quality of the generated data. The cultural background is the preprocessed text segment that contains language-specific knowledge. The dialogue history provides the historical questions and answers, which is set to an empty string when generating the initial question.

\begin{figure}[h]
\begin{tcolorbox}[colback=blue!2,colframe=blue!50!black]
\small
\textcolor{sblue}{\textbf{\{system prompt\}}}
\textcolor{sred}{\textbf{\{principles\}}} \\
\texttt{<document>}
\textcolor{sgreen}{\textbf{\{cultural background\}}}
\texttt{<\textbackslash document>}
\textcolor{orange}{\textbf{\{dialogue history\}}}
\end{tcolorbox}
\caption{Structure of the prompts used for dialogue generation. The provided cultural background is enclosed within a pair of separators.}
\label{fig:prompt-structure}
\end{figure}

\paragraph{Generating the Initial Dialogue} The principles used to generate the first question are shown in Figure~\ref{fig:first-question}. We ask the involved LLM (i.e., GPT-3.5) to understand the provided cultural background and then propose a related question that can be answered according to the cultural background.
For the generation of answers, we provide only a concise description of the principles in Figure~\ref{fig:first-answer} due to space limitations.
For each language, the principles are translated by humans into the target language. We only show the English version of the prompt to better understand the method.

\begin{figure}[h]
\begin{tcolorbox}[colback=red!2,colframe=red!50!black]
\small
\texttt{1. Pose "why" and "how" questions: given the provided document, ask why something happens or how it occurs. The questions should guide respondents to engage in more in-depth analysis and explanation, rather than simply stating facts. \\
2. Compare and contrast: if the text mentions a phenomenon or viewpoint, you can try comparing it with other similar situations and then pose questions to explore the similarities and differences between them, as well as potential impacts. \\
3. Predict future developments: if the text refers to a trend or direction of development, you can pose questions to discuss possible changes in the future or express opinions and predictions about a particular trend. \\
4. Stimulate reflection and discussion: Pose open-ended questions to encourage respondents to delve into deeper reflection and discussion.}
\end{tcolorbox}
\caption{Principles for generating the initial question.}
\label{fig:first-question}
\end{figure}

\begin{figure}[h]
\begin{tcolorbox}[colback=red!2,colframe=red!50!black]
\small
\texttt{1. Understand the content.\\
2. Logically reason about details.\\
3. Compare relevant situations.\\
4. Discuss future trends.\\
5. Engage in deeper discussion.}
\end{tcolorbox}
\caption{A brief description of the principles for generating the initial answer.}
\label{fig:first-answer}
\end{figure}

\paragraph{Generating Subsequent Dialogues}

After generating the initial question and answer, we iteratively produce subsequent dialogues. To improve the diversity of constructed dialogues, we propose two types of subsequent questions.
At each turn, we randomly decide whether to present an {\em in-depth question} for a more detailed exploration of the same topic or to generate an {\em expansive question} to delve into other subjects. The principles used to ask an in-depth question are shown in Figure~\ref{fig:depth-question}, while the principles used to ask an expansive question are shown in Figure~\ref{fig:width-question}. Note that when generating subsequent dialogues, the cultural background is also provided to the model. We will attach all the full prompts in supplementary materials.

\begin{figure}[ht]
\begin{tcolorbox}[colback=red!2,colframe=red!50!black]
\small
\texttt{1. Understand the context.\\
2. Uncover implicit information.\\
3. Challenge existing viewpoints.\\
4. Extend the topic.\\
5. Pose open-ended questions.\\
6. Delve into more complex logic.}
\end{tcolorbox}
\caption{A brief description of the principles to ask an in-depth following question.}
\label{fig:depth-question}
\end{figure}

\begin{figure}[h]
\begin{tcolorbox}[colback=red!2,colframe=red!50!black]
\small
\texttt{1. Abstract the theme.\\
2. Turn into overarching topics.\\
3. Considering temporal and spatial span.\\
4. Connect to related fields. \\
5. Take a global perspective.}
\end{tcolorbox}
\caption{A brief description of the principles to ask an expansive following question.}
\label{fig:width-question}
\end{figure}

Using the aforementioned approach, we automatically construct language-specific multi-turn conversations in four languages.
The details of constructed data will be illustrated in Section~\ref{sec: data details}, including the average length and some other statistics.
Note that the proposed knowledge-grounded data augmentation approach can also be applied to any other language.

\subsection{Language-Agnostic Data Crution}

In addition to language-specific abilities, the general abilities that are language-agnostic are also essential for LLMs. As numerous high-quality English SFT datasets already encompass a broad spectrum of general abilities, we suggest employing a two-stage translation mechanism to maximize the utility of existing English resources. Our goal is to reduce translation errors caused by cultural differences since some questions can not be directly translated into other languages (e.g., \texttt{write an English poem where each sentence starts with the letter ``A"}). In the first stage, we introduce a multi-criteria mechanism to filter out English-specific conversations that are difficult to translate accurately into other languages. Then we use GPT-3.5 to translate the remaining language-agnostic data. 
In this study, we consider three key components of general abilities for LLMs: chat, math reasoning, and code generation. For chat, we use ShareGPT as the English chat data, which consists of multi-turn dialogues between human users and ChatGPT. For math reasoning, we use MetaMath~\cite{metamath} as the English math data. For code generation, we use the Magicoder dataset~\cite{wei2023magicoder} as the English code data.

\subsubsection{Multi-Criteria Filter}

The criteria employed to filter out English-specific conversations are outlined in Figure~\ref{fig:filter_prompt}. Our goal is to retain only conversations whose topics can be discussed in any cultural background.
GPT-3.5 is utilized to ascertain whether a conversation contains information relevant to the specified features.
For instance, the conversations that include English jokes will be removed before translation.

\begin{figure}[h]
\begin{tcolorbox}[colback=blue!2,colframe=blue!50!black]
\small
\texttt{1. Full name of *human*.\\
2. Country, region, state, province, city, address.\\
3. Conventions, politics, history, and religion.\\
4. Poetry, rhymes, myths, tales, jokes, and slang.\\
5. Food, cloth, furniture, construction.\\
6. Organization, company, product, brand.}
\end{tcolorbox}
\caption{Criteria used to identify English-specific conversation. We only provide a brief version with a detailed explanation due to space limitations.}
\label{fig:filter_prompt}
\end{figure}

\subsubsection{Translator}

After the filtering process, the remaining conversations undergo the translation procedure, wherein they are translated into four languages using GPT-3.5-turbo to maintain fluency and accuracy. We also provide some translation principles to help GPT-3.5 better perform the translation, which is shown in Figure~\ref{fig:translate_prompt}.

\begin{figure}[h]
\begin{tcolorbox}[colback=blue!2,colframe=blue!50!black]
\small
\texttt{1. Ensure the completeness and consistency of content during the translation process, without adding or deleting any information.\\
2. Ensure that the translated text is fluent and natural, using the most common expressions in the target language whenever possible. Use officially prescribed translations for professional terms and adhere to the target-language expression conventions.\\
3. If certain terms are not in natural language but are mathematical symbols, programming languages, or LaTex language, please directly copy the original text.\\
4. If there are no equivalent translation terms for certain vocabulary, please directly copy the original text.\\
5. For citations and references, please directly copy the original text.}
\end{tcolorbox}
\caption{Translation principles.}
\label{fig:translate_prompt}
\end{figure}

\subsection{Data Pruning}
\label{sec:volume}

English math and code datasets are frequently extensive, exemplified by MetaMath~\cite{metamath} with 395K training examples and Magicoder~\cite{wei2023magicoder} comprising 186K training examples. Assuming the English data consists of $N$ training examples, the overall multilingual dataset would encompass $k \times N$ examples if we translate all the English training examples into other languages, where $k$ is the number of languages. The linear increase in data volume will result in higher training costs during SFT. As math and code problems are not closely tied to the cultural backgrounds of different countries, LLMs may have the capability to transfer English math and code abilities into other languages with only limited training examples. In other words, it may not be necessary for LLMs to learn all translated math and code problems.
To verify the assumption mentioned above, we conduct experiments on Chinese math and code tasks. For comparison, we fine-tune Llama-2-7b~\cite{llama2} in the following two different ways:
\begin{itemize}
    \item {\em From En SFT Model}: we first use English math or code data to fine-tune the base model, and then use different amounts of Chinese data to further tune the model.
    \item {\em From Base Model}: we directly use Chinese math or code data to fine-tune the base model.
\end{itemize}
\begin{figure}[h]
    \centering
    \includegraphics[width=0.99\linewidth]{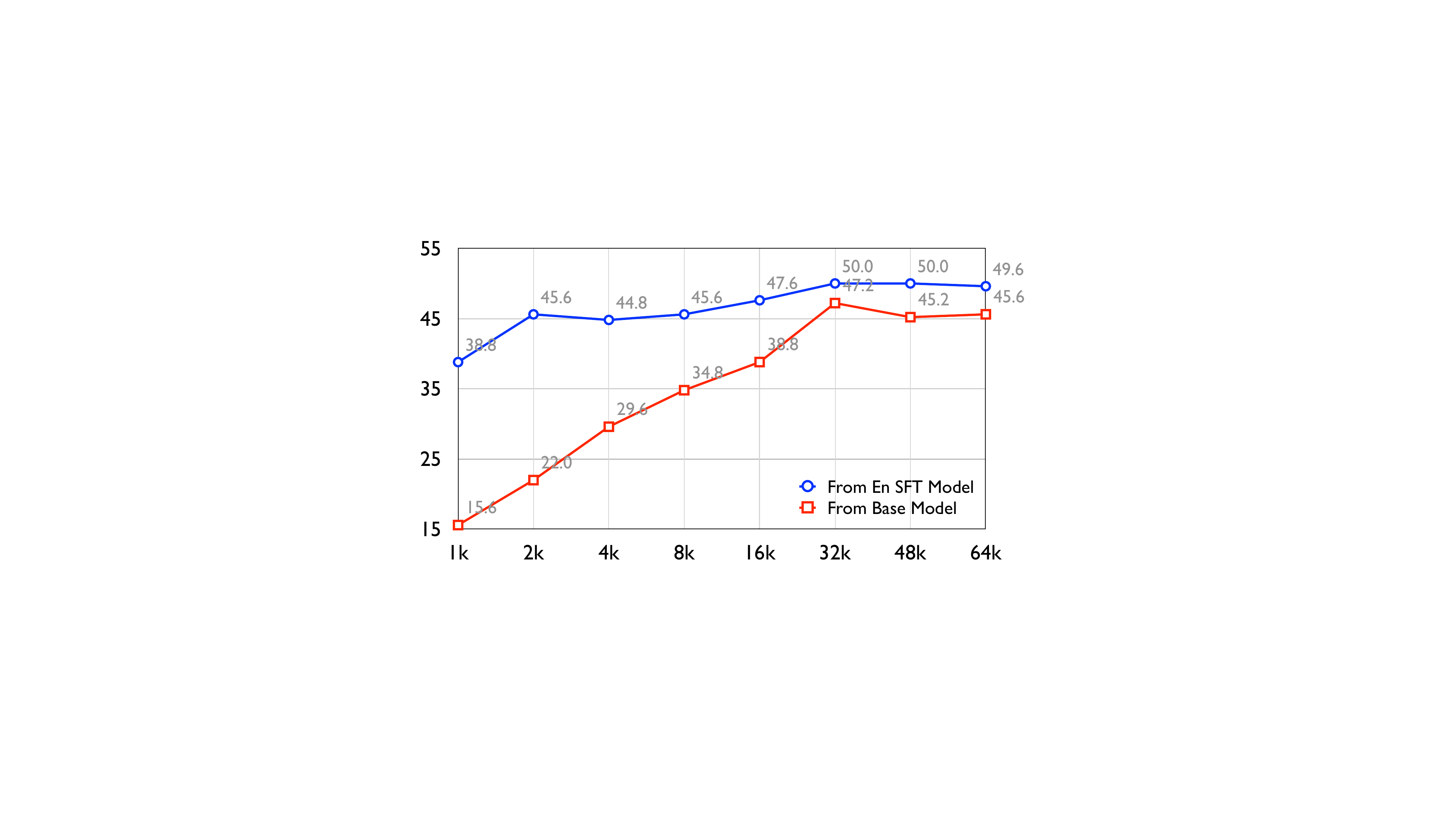}
    \caption{Performance on MGSM-Zh with different numbers of Chinese mathematical training examples.}
    \label{fig:math_data_volume}
\end{figure}
\begin{figure}[h]
    \centering
    \includegraphics[width=0.99\linewidth]{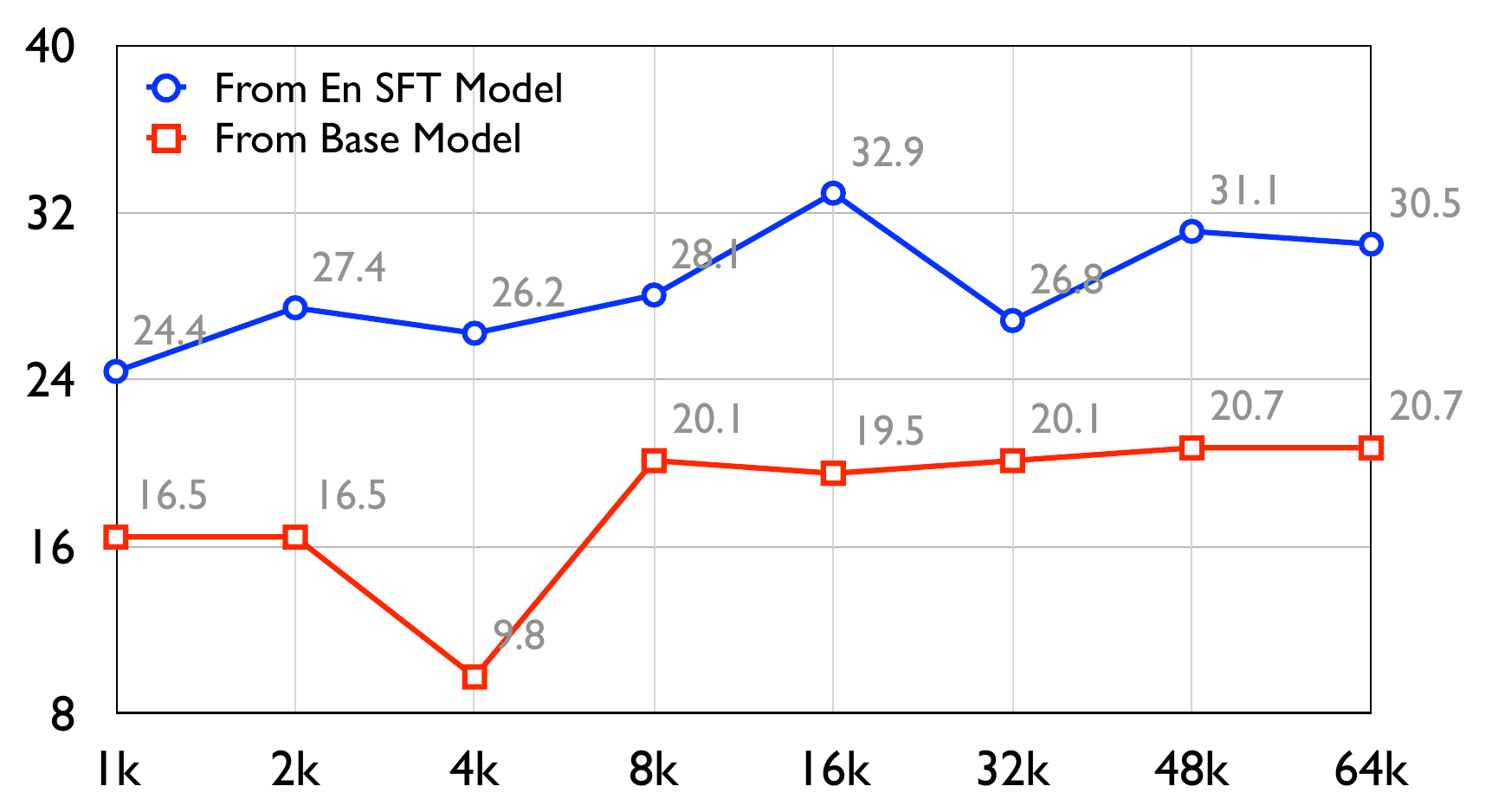}
    \caption{Performance on HumanEval-Zh with different numbers of Chinese code training examples.}
    \label{fig:code_data_volume}
\end{figure}
Figure~\ref{fig:math_data_volume} and \ref{fig:code_data_volume} show the performances of the two types of models. Surprisingly, the involved LLM exhibits strong cross-lingual transfer capabilities. For instance, utilizing only 2K Chinese mathematical training examples can yield a score of 45.6 when fine-tuning from the English SFT model. In contrast, directly fine-tuning the base model with an equivalent amount of Chinese data results in a significantly lower score of 22.0, highlighting the superior performance achieved through transfer from the English SFT model. In the Chinese code generation task, we observe a similar trend, wherein transfer learning from the English SFT model substantially enhances the performance of the model.

Moreover, we find that using more Chinese SFT data does not consistently lead to improved performance. For the math task, using 32K Chinese training examples achieves the best result. For the code task, the peak performance is attained with 16K Chinese code generation examples. Hence, we incorporate only 32K mathematical training examples and 16K code training examples for each non-English language in the UltraLink dataset.

\begin{table}[ht]
    \centering
    \small
    \begin{tabular}{l rrrr}
      \toprule
      \multirow{2}{*}{\bf Lang.} & \multicolumn{1}{c}{\bf Lang.Spec.} & \multicolumn{3}{c}{\bf Lang.Agno.} \\
      \cmidrule(lr){2-2} \cmidrule(lr){3-5}
      & \multicolumn{1}{c}{\bf Chat} & \multicolumn{1}{c}{\bf Chat} & \multicolumn{1}{c}{\bf Math} & \multicolumn{1}{c}{\bf Code} \\
      \midrule
      En & 10K & 67K & 395K & 186K \\
      \cmidrule(lr){1-1} \cmidrule(lr){2-5}
      Zh & 36K & 11K & 32K & 16K \\
      Ru & 37K & 11K & 32K & 16K \\
      Fr  & 30K & 11K & 32K & 16K \\
      Es & 34K & 11K & 32K & 16K \\
      \cmidrule(lr){1-1} \cmidrule(lr){2-5}
      UltraLink    & 147K & 112K & 523K & 250K \\
      ~w/o En & 137K & 45K & 128K & 64K \\
      \bottomrule
    \end{tabular}
    \caption{Scales of different components in UltraLink, which are measured by the number of dialogues.}
    \label{tab:datasets_quantity}
\end{table}

\begin{table*}[ht]
\setlength{\tabcolsep}{13.5pt}
\centering
\small
\begin{tabular}{l rrrrr}
  \toprule
  \multirow{2}{*}{\bf Dataset} & \multirow{2}{*}{\bf Dialogues} & \multirow{2}{*}{\bf Turns} & \multicolumn{3}{c}{\bf Average Length} \\
  \cmidrule(lr){4-6}
  & & & \bf Question & \bf Answer & \bf Turn \\
  \midrule
  Okapi Dataset~\cite{okapi} & 207K & 207K & 28.64 & 95.72 & 124.36\\
  Guanaco Dataset~\cite{Guanaco} & \bf 1173K & 1173K & 77.58 & 83.31 & 160.89\\
  Multialpaca~\cite{polylm} & 132K & 132K & 39.86 & 83.71 & 123.57\\
  Phoenix SFT data~\cite{phoenix} & 464K & 893K & \bf 165.27 & 200.07 & 365.34\\
  \midrule
  UltraLink~(\textbf{Ours}) & 1032K & \bf 1623K & 87.86 & \bf 290.35 & \bf 378.21 \\
  \bottomrule
\end{tabular}
\caption{Comparison between UltraLink and existing open-source multilingual SFT datasets.}
\label{tab:datasets_comparison}
\end{table*}

\begin{table*}[t]
  \setlength{\tabcolsep}{8pt}
  \centering
  \small
  \begin{tabular}{l l l rrrrrr }
    \toprule
    \multirow{2}{*}{\bf Model} &  \multirow{2}{*}{\bf Backbone} & \multirow{2}{*}{\bf SFT Data} & \multicolumn{6}{c}{{\bf OMGEval (Chat)}}  \\
    \cmidrule(lr){4-9}
    & & & En & Zh & Es & Ru & Fr & Avg. \\
    \midrule
     Bloomz-7b1-mt & Bloomz-7b1 & xP3mt & 0.0 & 0.9 & 0.1 & 0.5 & 0.3 & 0.4 \\
     Phoenix-inst-chat-7b & Bloomz-7b1 & Phoenix SFT data & 6.9 & 13.3 & 7.4 & 2.9 & 8.1 & 7.7 \\
     PolyLM-Multialpaca-13b & PolyLM-13b & Multialpaca & 3.4 & 5.0 & 2.1 & 5.1 & 2.2 & 3.6 \\
     PolyLM-Chat-13b & PolyLM-13b & Closed-source & 7.7 & 14.0 & 6.1 & 5.5 & 4.8 & 7.6 \\
     Chimera-inst-chat-13b & Llama-13b & Phoenix SFT data & 15.5 & 9.7 & 11.8 & 13.7 & 13.8 & 12.9 \\
     Okapi-7b &  Llama-2-7b &  Okapi Dataset & 8.8 & 6.2 & 5.0 & 12.1 & 8.7 & 8.2 \\
     Guanaco-7b & Llama-2-7b & Guanaco Dataset & 4.6 & 3.8 & 0.4 & 1.8 & 1.2 & 2.4 \\
     Guanaco-13b & Llama-2-13b & Guanaco Dataset & \bf 29.0 & 8.6 & 16.9 & 15.4 & 17.3 & 17.5 \\
     \cmidrule(lr){1-1} \cmidrule(lr){2-2} \cmidrule(lr){3-3} \cmidrule(lr){4-9}  
     UltraLink-LM & Llama-2-13b & UltraLink &  28.8 & \bf 21.9 & \bf 23.5 & \bf 37.6 & \bf 29.0 & \bf 28.2  \\
     \midrule
     \multirow{2}{*}{\bf Model} &  \multirow{2}{*}{\bf Backbone} & \multirow{2}{*}{\bf SFT Data} & \multicolumn{6}{c}{{\bf Multilingual HumanEval (Code)}}  \\
    \cmidrule(lr){4-9}
    & & & En & Zh & Es & Ru & Fr & Avg. \\
    \midrule
     Bloomz-7b1-mt & Bloomz-7b1 & xP3mt & 8.5 & 7.3 & 6.1 & 8.5 & 6.1 & 7.3 \\
     Phoenix-inst-chat-7b & Bloomz-7b1 & Phoenix SFT data & 11.0 & 10.4 & 8.5 & 1.2 & 13.4 & 12.2 \\
     PolyLM-Multialpaca-13b & PolyLM-13b & Multialpaca & 8.5 & 7.3 & 6.1 & 6.1 & 6.1 & 6.8 \\
     PolyLM-Chat-13b & PolyLM-13b & Closed-source & 10.4 & 7.9 & 6.1 & 7.3 & 8.5 & 8.1 \\
     Chimera-inst-chat-13b & Llama-13b & Phoenix SFT data & 14.6 & 13.4 & 14.6 & 12.8 & 14.0 & 13.9 \\
     Okapi-7b &  Llama-2-7b &  Okapi Dataset & 12.2 & 11.0 & 8.5 & 8.5 & 8.5 & 9.8 \\
     Guanaco-7b & Llama-2-7b & Guanaco Dataset & 9.2 & 6.7 & 11.0 & 9.8 & 12.8 & 9.9 \\
     Guanaco-13b & Llama-2-13b & Guanaco Dataset & 18.3 & 15.9 & 9.8 & 8.5 & 14.6 & 12.2 \\
     \cmidrule(lr){1-1} \cmidrule(lr){2-2} \cmidrule(lr){3-3} \cmidrule(lr){4-9} 
     UltraLink-LM & Llama-2-13b & UltraLink & \bf 60.4 & \bf 43.9 & \bf 40.9 & \bf 49.4 & \bf 39.6 & \bf 46.8  \\
    \midrule
        \multirow{2}{*}{\bf Model} &  \multirow{2}{*}{\bf Backbone} & \multirow{2}{*}{\bf SFT Data} & \multicolumn{6}{c}{{\bf MGSM (Math)}}  \\
    \cmidrule(lr){4-9}
    & & & En & Zh & Es & Ru & Fr & Avg. \\
    \midrule
     Bloomz-7b1-mt & Bloomz-7b1 & xP3mt & 2.8 & 1.6 & 2.0 & 0.4 & 2.8 & 1.7 \\
     Phoenix-inst-chat-7b & Bloomz-7b1 & Phoenix SFT data & 3.2 & 3.2 & 2.8 & 3.2 & 3.2 & 3.1 \\
     PolyLM-Multialpaca-13b & PolyLM-13b & Multialpaca & 1.2 & 2.8 & 1.6 & 2.8 & 2.4 & 2.4 \\
     PolyLM-Chat-13b & PolyLM-13b & Closed-source & 10.8 & 6.4 & 4.8 & 4.4 & 5.6 & 5.3 \\
     Chimera-inst-chat-13b & Llama-13b & Phoenix SFT data & 14.0 & 11.6 & 10.0 & 12.0 & 12.8 & 11.6 \\
     Okapi-7b &  Llama-2-7b &  Okapi Dataset & 4.0 & 2.4 & 3.6 & 4.4 & 4.8 & 3.8 \\
     Guanaco-7b & Llama-2-7b & Guanaco Dataset & 4.0 & 1.6 & 3.2 & 2.8 & 4.4 & 3.0 \\
     Guanaco-13b & Llama-2-13b & Guanaco Dataset & 13.6 & 10.8 & 11.2 & 6.4 & 5.2 & 8.4 \\
     \cmidrule(lr){1-1} \cmidrule(lr){2-2} \cmidrule(lr){3-3} \cmidrule(lr){4-9} 
     UltraLink-LM & Llama-2-13b & UltraLink & \bf 70.4 & \bf 56.0 & \bf 70.4 & \bf 64.8 & \bf 63.6 & \bf 63.7  \\
    \bottomrule
  \end{tabular}
\caption{Performance of the involved multilingual SFT LLMs on different tasks.}
\label{tab: LLM_scores}
\end{table*}

\section{Dataset Statistics}
\label{sec: data details}
\subsection{Data Distribution}

Table~\ref{tab:datasets_quantity} presents the scale of each component in UltraLink, encompassing five languages. Each language contributes four types of SFT data: chat data with language-specific knowledge, chat data with language-agnostic knowledge, math data, and code data. The quantities of language-agnostic segments are approximately equal for the four non-English languages.

\subsection{Comparison with Existing Datasets}

Before us, there are some existing multilingual SFT datasets, where we select four representative datasets for comparison, including the Okapi dataset~\cite{okapi}, the Guanaco dataset~\cite{Guanaco}, Multialpaca~\cite{polylm}, and the Phoenix SFT data~\cite{phoenix}. We conduct a comparison based on the number of dialogues, the number of conversation turns, and the average lengths across the respective datasets. As shown in Table~\ref{tab:datasets_comparison}, we find that UltraLink contains fewer dialogues than the Guanaco dataset, but the latter only contains single-turn conversations. Only the Phoenix SFT data and UltraLink include multi-turn conversations.

We use the number of tokens estimated by \texttt{tiktoken} as the length for each question and answer. The question token length does not include the document. On average, UltraLink exhibits the longest average length per turn (i.e., 378.21 tokens), considering both questions and their corresponding answers. Compared to UltraLink, the Phoenix SFT data has longer questions (165.27 vs. 87.86), but its answers are shorter (200.07 vs. 290.35).

\begin{figure}[ht]
    \centering
    \includegraphics[width=0.99\linewidth]{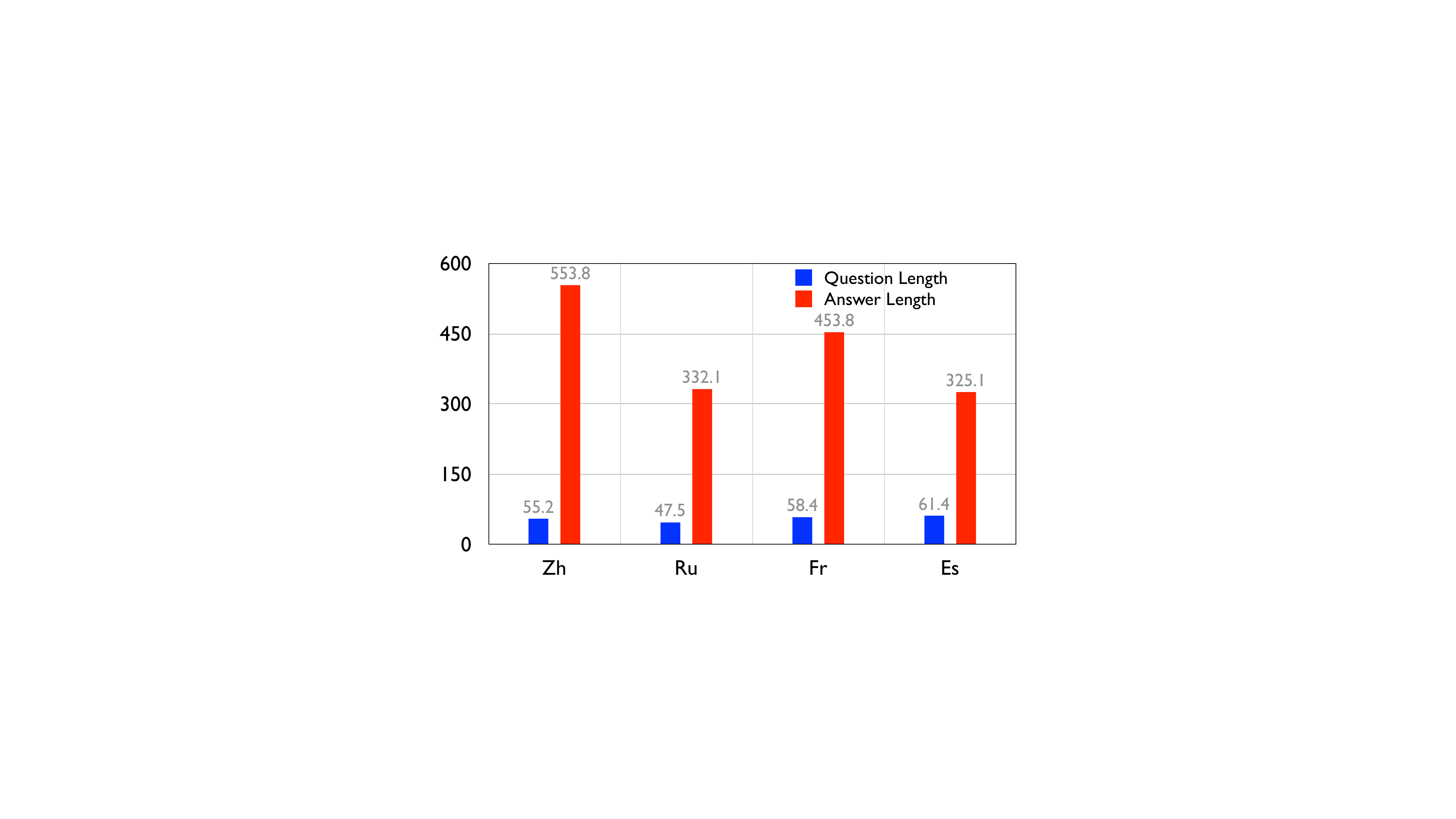}
    \caption{Number of tokens for each language.}
    \label{fig:token_len_5lang}
\end{figure}

For each language, we also estimate the average lengths of questions and answers, and the results are shown in Figure~\ref{fig:token_len_5lang}. Across all languages, the answer is significantly longer than the question.

\section{Experiment}
\subsection{Setup}
\paragraph{Baselines} For thorough comparison, we select several representative multilingual baselines in our experiments, including Bloomz-7b1-mt~\cite{bloom}, Phoenix-inst-chat-7b~\cite{phoenix}, PolyLM-Multialpaca-13b~\cite{polylm}, PolyLM-Chat-13b~\cite{polylm}, Chimera-inst-chat-13b~\cite{phoenix}, Okapi-7b~\cite{okapi}, Guanaco-7b~\cite{Guanaco}, and Guanaco-13b~\cite{Guanaco}.
Okapi-7b is fine-tuned by ourselves based on Llama-2-7b using the Okapi dataset, while other baselines are downloaded from Huggingface\footnote{\url{https://huggingface.co}}.

\paragraph{Training details} Based on Llama-2-13b~\cite{llama}, UltraLink-LM is fine-tuned with the constructed UltraLink dataset for 3 epochs. We use the cosine learning rate schedule and the peak learning rate is set to 2e-5. The warm-up ratio is set to 0.04. Each mini-batch contains 128 training examples in total. The maximum sequence length is 4096. We train the model using 32 A100 GPUs for about 140 hours.

\paragraph{Evaluation} We examine the model performance on three tasks, including chat, math reasoning, and code generation. For chat, we use OMGEval~\cite{OMGEval} for evaluation, which is a multilingual version of the widely-used English benchmark  AlpacaEval~\cite{alpaca_eval}.
OMGEval is not a mere translated version of AlpacaEval. Instead, it localizes the English questions according to the cultural backgrounds of each language.
We employ MGSM~\cite{mgsm} to evaluate math reasoning abilities, which is also a multilingual benchmark.
Since there are no existing multilingual test sets for code generation, we use GPT-3.5 with carefully designed prompts to translate HumanEval~\cite{humaneval} into other languages, which serves as the multilingual benchmark to evaluate the code abilities of LLMs. We use the \texttt{UltraEval} toolkit\footnote{\url{https://github.com/OpenBMB/UltraEval}} for model inference and evaluation, which supports a wide range of open-source models.

\subsection{Results}

Table~\ref{tab: LLM_scores} shows the results of the involved multilingual SFT LLMs on different tasks. In terms of general chat abilities, our model achieves the best average results. While Guanaco-13b slightly outperforms us in English (29.0 vs. 28.8), its performance is notably lower than ours in non-English languages. Given that Guanaco-13b shares the same backbone (i.e., Llama-2-13b) with our model, the results imply the superiority of the proposed UltraLink dataset.

For the code generation Task, previous multilingual SFT datasets did not take into account the multilingual code abilities, which we think is very important in many real-world scenarios. Our model achieves a score of 60.4 in the English HumanEval benchmark, surpassing even CodeLlama-34b-Python~\cite{codellama}. For comparison, training the model solely on the English Magicoder~\cite{wei2023magicoder} dataset results in a HumanEval score of 53.0. The improvement of UltraLink-LM over the model trained on the English Magicoder dataset (i.e., 60.4 vs. 53.0) suggests that the constructed code SFT data in other languages can also enhance English code abilities. This confirms our assumption that modern LLMs possess strong transfer abilities for language-agnostic skills.

In the math reasoning task, our model consistently outperforms all other baselines across all five languages. The performance of UltraLink-LM in both math and code tasks underscores the effectiveness of our method in enabling multilingual LLMs to acquire general abilities.

\section{Related Works}
\paragraph{Supervised Fine-tuning} SFT is now a crucial part of constructing a powerful LLM. SODA~\cite{kim-etal-2023-soda} constructs high-quality social dialogues by contextualizing social commonsense knowledge from a knowledge graph. Using the technique of self-instruct~\cite{wang-etal-2023-self-instruct}, Alpaca~\cite{alpaca} is one of the pioneers to leverage ChatGPT to collect SFT data. 
UltraChat~\cite{ding-etal-2023-enhancing} utilizes ChatGPT to generate topics in a tree-style structure for the construction of large-scale dialogues. With these efforts, English SFT resources are becoming increasingly rich and effective.

\paragraph{Multilingual SFT Datasets}
To enhance the global utility of LLMs, numerous multilingual SFT datasets have been created. \citet{okapi} employ ChatGPT to translate Alpaca into various languages. \citet{phoenix} combine ShareGPT with Alpaca and then translate the two datasets. \citet{Guanaco} and \citet{polylm} extend tasks from Alpaca by introducing filters and rewrites of seed tasks in different languages, generating datasets through multiple iterations. This work proposes the utilization of a multilingual knowledge base to enhance the cultural diversity of multilingual Supervised Fine-Tuning data, as well as to improve the language-agnostic general abilities of LLMs through cross-lingual transfer learning.

\section{Conclusion}
In this work, we propose a knowledge-grounded data augmentation method and a two-stage translation mechanism to construct language-specific and language-agnostic multilingual SFT data, respectively.
Experiments demonstrate that the proposed dataset is effective for multilingual LLMs.

\section{Ethical Impact}
We present a framework for generating SFT data across diverse languages and use the proposed dataset to learn an LLM. Our LLM may inevitably encounter common challenges, including issues such as hallucination and toxicity. We highly recommend users utilize our work exclusively for research purposes, to enhance the efficacy of LLMs across various languages.

\section{Limitations}
In the paper, our proposed data construction framework is only applied to four language types. Nevertheless, the framework can be easily extended to other languages. We leave it to the future work to include more languages.
Moreover, due to constraints imposed by the base model, the multilingual capability still faces several limitations. Notably, the model exhibits significantly better performance in English across many tasks. There is a pressing need to continue constructing high-quality pre-training multilingual datasets, to unlock the full potential of multilingual abilities in LLMs.

\bibliography{custom}

\end{document}